# A robust, low-cost approach to Face Detection and Face Recognition

Divya Jyoti[1], Aman Chadha[2], Pallavi Vaidya[3], and M. Mani Roja[4]

*Abstract*— In the domain of Biometrics, recognition systems based on iris, fingerprint or palm print scans etc. are often considered more dependable due to extremely low variance in the properties of these entities with respect to time. However, over the last decade data processing capability of computers has increased manifold, which has made real-time video content analysis possible. This shows that the need of the hour is a robust and highly automated Face Detection and Recognition algorithm with credible accuracy rate. The proposed Face Detection and Recognition system using Discrete Wavelet Transform (DWT) accepts face frames as input from a database containing images from low cost devices such as VGA cameras, webcams or even CCTV's, where image quality is inferior. Face region is then detected using properties of L*a*b* color space and only Frontal Face is extracted such that all additional background is eliminated. Further, this extracted image is converted to grayscale and its dimensions are resized to 128 x 128 pixels. DWT is then applied to entire image to obtain the coefficients. Recognition is carried out by comparison of the DWT coefficients belonging to the test image with those of the registered reference image. On comparison, Euclidean distance classifier is deployed to validate the test image from the database. Accuracy for various levels of DWT Decomposition is obtained and hence, compared.

## I. INTRODUCTION

A face recognition system is essentially an application [1] intended to identify or verify a person either from a digital image or a video frame obtained from a video source. Although other reliable methods of biometric personal identification exist, for e.g., fingerprint analysis or iris scans, these methods inherently rely on the cooperation of the participants, whereas a personal identification system based on analysis of frontal or profile images of the face is often effective without the participant's cooperation or intervention. Automatic identification or verification may be achieved by comparing selected facial features from the image and a facial database. This technique is typically used in security systems. Given a large database of images and a photograph, the problem is to select from the database a small set of records such that one of the image records matched the photograph. The success of the method could be measured in terms of the ratio of the answer list to the number of records in the database. The recognition problem is made difficult by the great variability in head rotation and tilt, lighting intensity and angle, facial expression, aging, etc. A robust facial recognition system must be able to cope with the above factors and yet provide satisfactory accuracy levels. A general statement of the problem of machine recognition of faces can [2] be formulated as: given a still or video image of a scene, identify or verify one or more persons in the scene using a stored database of faces. The solution to the problem involves segmentation of faces, feature extraction from face regions, recognition, or verification. In identification problems, the input to the system is an unknown face, and the system reports back the determined identity from a database of known individuals, whereas in verification problems, the system needs to confirm or reject the claimed identity of the input face.

Some of the various applications of face recognition include driving licenses, immigration, national ID, passport, voter registration, security application, medical records, personal device logon, desktop logon, human-robot-interaction, human-computer-interaction, smart cards etc. Face recognition is such a challenging yet interesting problem that it has attracted researchers who have different backgrounds: pattern recognition, neural networks, computer vision, and computer graphics, hence the literature is vast and diverse. The usage of a mixture of techniques makes it difficult to classify these systems based on what types of techniques they use for feature representation or classification. To have clear categorization, the proposed paper follows the holistic approach [2]. Specifically, the following techniques are employed for facial feature extraction and recognition:

1) Holistic matching methods: These methods use the whole face region as a raw input to the recognition system. One of the most widely used representations of the face region is Eigenpictures, which is inherently based on principal component analysis.
2) Feature-based matching methods: Generally, in these methods, local features such as the eyes, nose and mouth are first extracted and their locations and local statistics are fed as inputs into a classifier.
3) Hybrid methods: It uses both local features and whole face region to recognize a face. This method could potentially offer the better of the two types of methods.

Manuscript received September 11, 2011.
[1] D. J. Rajdev is with the Thadomal Shahani Engineering College, Mumbai, 400002, INDIA (phone: +91-8879100684; e-mail: dj.rajdev@gmail.com).
[2] A. R. Chadha is with the Thadomal Shahani Engineering College, Mumbai, 400002, INDIA (phone: +91-9930556583; e-mail: aman.x64@gmail.com).
[3] P. P. Vaidya is with the Thadomal Shahani Engineering College, Mumbai, 400002, INDIA (e-mail: pallavi.p.vaidya@gmail.com).
[4] M. M. Roja is an Associate Professor in the Electronics and Telecommunication Engineering Department, Thadomal Shahani Engineering College, 400050, INDIA (e-mail: maniroja@yahoo.com).

Most electronic imaging applications often desire and require high resolution images. 'High resolution' basically means that pixel density within an image is high, and therefore a HR image can offer more details and subtle transitions that may be critical in various applications [19]. For instance, high resolution medical images could be very helpful for a doctor to make an accurate diagnosis. It may be easy to distinguish an object from similar ones using high resolution satellite images, and the performance of pattern recognition in computer vision can easily be improved if such images are provided. Over the past few decades, charge-coupled device (CCD) and CMOS image sensors have been widely used to capture digital images. Although these sensors are suitable for most imaging applications, the current resolution level and consumer price will not satisfy the future demand [19].

Past studies by researches and scientists that have investigated the challenging task of face detection and recognition have therefore, typically used high resolution images. Moreover, most standard face databases such as the MIT-CBCL Face Recognition Database [21], CMU Multi-PIE [22], The Yale Face Database [23] etc., that are basically used as a standard test data set by researchers to benchmark their results, also employ high quality images.

Results obtained by solutions proposed by researchers are therefore, relevant for theoretical understanding of face detection and identification in most cases. Practical conditions being rarely optimal, a number of factors play an important role in hampering system performance. Image degradation, i.e., loss of resolution caused mainly by large viewing distances as demonstrated in [4], and lack of specialized high resolution image capturing equipment such as commercial cameras are the underlying factors for poor performance of face detection and recognition systems in practical situations. There are two paradigms to alleviate this problem, but both have clear disadvantages. One option is to use super-resolution algorithms to enhance the image as proposed in [20], but as resolution decreases, super-resolution becomes more vulnerable to environmental variations, and it introduces distortions that affect recognition performance. A detailed analysis of super-resolution constraints has been presented in [3]. On the other hand, it is also possible to match in the low-resolution domain by downsampling the training set, but this is undesirable because features important for recognition depend on high frequency details that are erased by downsampling. These features are permanently lost upon performing downsampling and cannot be recovered with upsampling [24].

The proposed system has been designed keeping in view these critical factors and to address such bottlenecks.

## II. IDEA OF THE PROPOSED SOLUTION

The database consists of a set of face samples of 50 people. There are 5 test images and 5 training or reference images. Frontal face images are detected and hence, extracted. DWT is applied to the entire image so as to obtain the global features which include approximate coefficients (low frequency coefficients) and detail coefficients (high frequency coefficients). The approximate coefficients thus obtained, are stored and the detail coefficients are discarded. Various levels of DWT are realized and their corresponding accuracy rates are determined.

### A. Frontal Face Image Detection and Extraction

The face detection problem can be defined as, given an input an arbitrary image, which could be a digitized video signal or a scanned photograph, determine whether or not there are any human faces in the image and if there are, then return a code corresponding to their location. Face detection as a computer vision task has many applications. It has direct relevance to the face recognition problem, because the first and foremost important step of an automatic human face recognition system is usually identifying and locating the faces in an unknown image [5].

For our purpose, face detection is actually a face localization problem in which the image position of single face has to be determined [6]. The goal of our facial feature detection is to detect the presence of features, such as eyes, nose, nostrils, eyebrow, mouth, lips, ears, etc., with the assumption that there is only one face in an image [7]. The system should also be robust against human affective states of like happy, sad, disgusted etc. The difficulties associated with face detection systems due to the variations in image appearance such as pose, scale, image rotation and orientation, illumination and facial expression make face detection a difficult pattern recognition problem. Hence, for face detection following problems need to be taken into account [5]:

1) Size: A face detector should be able to detect faces in different sizes. Thus, the scaling factor between the reference and the face image under test, needs to be given due consideration.
2) Expressions: The appearance of a face changes considerably for different facial expressions and thus, makes face detection more difficult.
3) Pose variation: Face images vary due to relative camera-face pose and some facial features such as an eye or the nose may become partially or wholly occluded. Another source of variation is the distance of the face from the camera, changes in which can result in perspective distortion.
4) Lighting and texture variation: Changes in the light source in particular can change a face's appearance can also cause a change in its apparent texture.
5) Presence or absence of structural components: Facial features such as beards, moustaches and glasses may or may not be present. And also there may be variability among these components including shape, colour and size.

The proposed system employs global feature matching for face recognition. However, all computation takes place only on the frontal face, by eliminating the hair and background as these may vary from one image to another. All systems therefore need frontal face extraction. One approach to achieve the aforementioned is by manually cropping the test image for required region or by precisely aligning the user's face with the camera before the test sample is clicked. Both these methods may introduce a high degree of human error and so they have been avoided. Instead, automated Frontal Face Detection and Extraction is put to use. Therefore, a robust automatic face recognition system should be capable of handling the above problem with no need for human intervention. Thus, it is practical and advantageous to realize automatic face detection in a functional face recognition system. Commonly used methods for skin detection include Gabor filters, neural networks and template matching.

It has been proved that Gabor filters give optimum output for a wide range of variations in the test image with respect to user image, but it is the most time intensive procedure [8]. Moreover, it is unlikely that the test image would be severely out of sync for on the spot face recognition, so this method is not used. Most neural-network based algorithms [18],[19] are tedious and require training samples for different skin types which add to the already vast reference image database; hence, even this does not fit the program's requirements. Even template matching has severe drawbacks, including high computational cost [10] and fails to work as expected when the user's face is positioned at an angle in the test image.

After considering all the above factors, a classical appearance based methodology is applied to extract Frontal face. The default sRGB colour space is transformed to L*a*b* gamut, because L*a*b* separates intensity from *a* and *b* colour components [11]. L*a*b* colour is designed to approximate human vision in contrast to the RGB and CMYK colour models. It aspires to perceptual uniformity, and its *L* component closely matches human perception of lightness. It can thus be used to make accurate colour balance corrections by modifying output curves in the *a* and *b* components, or to adjust the lightness contrast using the *L* component. In RGB or CMYK spaces, which model the output of physical devices rather than human visual perception, these transformations can only be done with the help of appropriate blend modes in the editing application [10]. This distinction makes L*a*b* space more perceptually uniform as compared to sRGB and thus identifying tones, and not just a single colour, can be accomplished using L*a*b* space.

Fig. 1 shows the RGB colour model (B) relating to the CMYK model (C). The larger gamut of L*a*b* (A) gives more of a spectrum to work with, thus making the gamut of the device the only limitation.

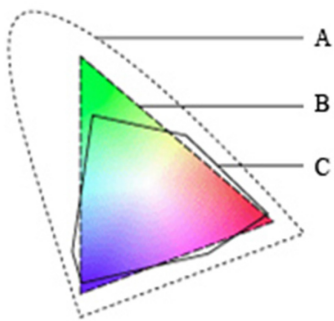

Fig. 1. RGB, CMYK and L*a*b* Colour Model

Tone identification is applied to detect skin by calculating the gray threshold of *a* and *b* colour components and then converting the image to pure black and white (BW) using the obtained threshold. Thus, RGB colour space can separate out only specific pigments, but L*a*b* space can separate out tones. Fig. 2 shows skin color differentiation in the form of white color using L*a*b* space whereas Fig. 4 shows no such differentiation using RGB. The extracted frontal face has been shown in Fig. 3.

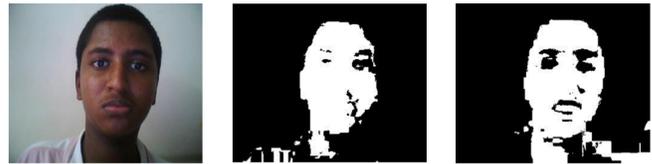

Fig. 2. Reference image (left), image in black and white *a* plane (middle) and image in black and white *b* plane (right)

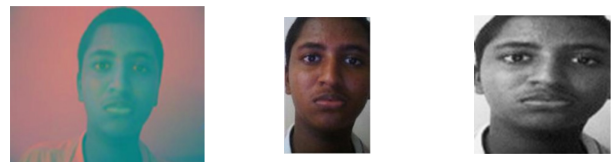

Fig. 3. Image in L*a*b* color space (left), frontal face extracted image (middle) and grayscale resized image (right)

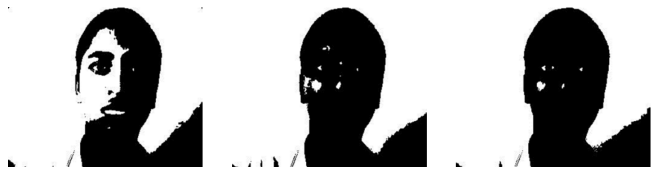

Fig. 4. B&W image of red color space (left), B&W image of green color space (middle) and B&W image of blue color space (right)

There is higher probability of skin surface being the lighter part of the image as compared to the gray threshold [12], This may happen due to illumination and natural skin colour (in most cultures), so, pure white regions in the black and white image correspond to skin. It is assumed that face will have at least one hole, i.e., a small patch of absolute black due to eyes, chin, dimples etc. [11] and on the basis of presence of holes frontal face is separated from other skin surfaces like hands. A bounding box is created around the Frontal Face and after cropping the excess area, frontal face extraction is complete.

The above technique has been tested extensively on images obtained from standalone VGA cameras, webcams and camera equipped mobile devices having a resolution of 640 × 480. Even for resolutions as low as 320 × 200, where the test image is poorly illuminated or extremely grainy, the algorithm was able to successfully extract frontal face from test images. Thus, the proposed system is robust enough to achieve desired result even when low cost equipment like CCTV's and low resolution webcams are used. Also, since equipment with inferior picture quality like CCTV's and low resolution webcams are used, the algorithm works as expected and hence can be called a low-cost approach to face detection. The extracted frontal face image is then fed as an input to the DWT-based face recognition process.

*B. Normalization*

Since the facial images are captured at different instants of the day or on different days, the intensity for each image may exhibit variations. To avoid these light intensity variations, the test images are normalized so as to have an average intensity value with respect to the registered image. The average intensity value of the registered images is calculated as summation of all pixel values divided by the total number of pixels. Similarly, average intensity value of the test image is calculated. The normalization value is calculated as:

$$\text{Normalization Value} = \frac{\text{Average value of reference image}}{\text{Average value of test image}} \quad (1)$$

This value is multiplied with each pixel of the test image. Thus we get a normalized image having an average intensity with respect to that of the registered image. Fig. 5 shows the test image and the corresponding normalized image.

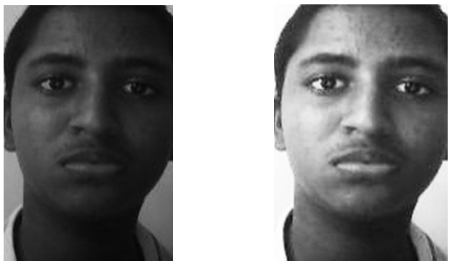

Fig. 5. Test Image and Normalized Image

*C. Discrete Wavelet Transform*

DWT [13] is a transform which provides the time-frequency representation. Often a particular spectral component occurring at any instant is of particular interest [14]. In these cases it may be very beneficial to know the time intervals these particular spectral components occur. For example, in EEGs, the latency of an event-related potential is of particular interest. DWT is capable of providing the time and frequency information simultaneously, hence giving a time-frequency representation of the signal. In numerical analysis and functional analysis, DWT is any wavelet transform for which the wavelets are discretely sampled. In DWT, an image can be analyzed by passing it through an analysis filter bank followed by decimation operation. The analysis filter consists of a low pass and high pass filter at each decomposition stage. When the signal passes through filters, it splits into two bands. The low pass filter which corresponds to an averaging operation, extracts the coarse information of the signal. The high pass filter which corresponds to a differencing operation, extracts the detail information of the signal. Fig. 6 shows the filtering operation of DWT.

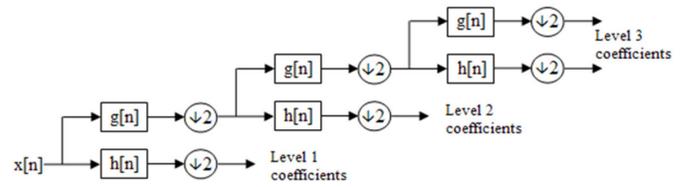

Fig. 6. DWT filtering operation

A two dimensional transform is accomplished by performing two separate one dimensional transforms. First the image is filtered along the row and decimated by two. It is then followed by filtering the sub image along the column and decimated by two. This operation splits the image into four bands namely LL, LH, HL and HH respectively. Further decompositions can be achieved by acting upon the LL sub band successively and the resultant image is split into multiple bands. For representational purpose, Level 2 decomposition of the normalized test image Fig. 5, is shown in Fig. 7.

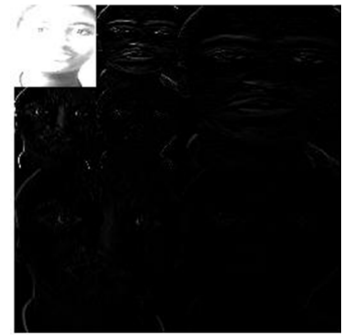

Fig. 7. Level 2 DWT decomposition

At each level in the above diagram, the frontal face image is decomposed into low and high frequencies. Due to the decomposition process, the input signal must be a multiple of 2n where n is the number of levels. The size of the input image at different levels of decomposition is illustrated in Fig. 8.

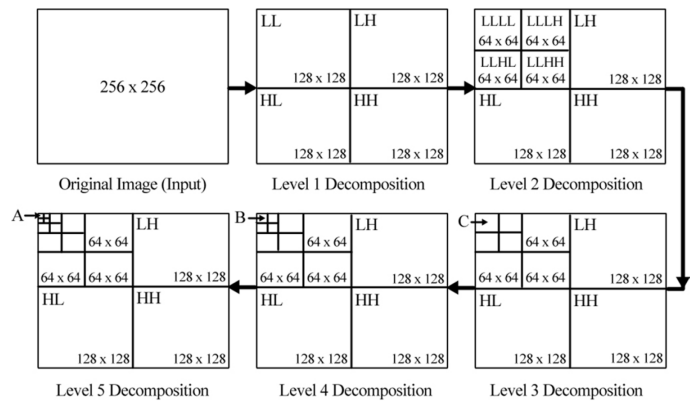

A: Block of size 8 x 8 ; B: Block of size 16 x 16 ; C: Block of size 32 x 32

Fig. 8. Size of the image at different levels of DWT decomposition

The first DWT was invented by the Hungarian mathematician Alfréd Haar. The Haar wavelet [15] is the first known wavelet and was proposed in 1909 by Alfred Haar. The term wavelet was coined much later. The Haar wavelet is also the simplest possible wavelet. Wavelets are mathematical functions developed for the purpose of sorting data by frequency. Translated data can then be sorted at a resolution which matches its scale. Studying data at different levels allows for the development of a more complete picture. Both small features and large features are discernable because they are studied separately. Unlike the Discrete Cosine Transform (DCT), the wavelet transform is not Fourier-based and hence, does a better job of handling discontinuities in data [16].

For an input represented by a list of 2n numbers, the Haar wavelet transform may be considered to simply pair up input values, storing the difference and passing the sum. This process is repeated recursively, pairing up the sums to provide the next scale, finally resulting in 2n − 1 differences and one final sum. Each step in the forward Haar transform calculates a set of wavelet coefficients and a set of averages. If a data set $s_0$, $s_1$, $s_{N-1}$ contains N elements; there will be N/2 averages and N/2 coefficient values. The averages are stored in the lower half of the N element array and the coefficients are stored in the upper half. The averages become the input for the next step in the wavelet calculation, where for iteration i+1, $N_{i+1} = N_i/2$. The Haar wavelet operates on data by calculating the sums and differences of adjacent elements. The Haar equations to calculate an average ($a_i$) and a wavelet coefficient ($c_i$) from an odd and even element in the data set can be given as:

$$a_i = \frac{(S_i + S_{i+1})}{2} \quad (2)$$

$$c_i = \frac{(S_i - S_{i+1})}{2} \quad (3)$$

In wavelet terminology, the Haar average is calculated by the scaling function while the coefficient is calculated by the wavelet function.

*D. Inverse Discrete Wavelet Transform*

The data input to the forward transform can be perfectly reconstructed using the following equations:

$$S_i = a_i + c_i \quad (4)$$

$$S_i = a_i - c_i \quad (5)$$

After applying DWT, we take approximate coefficients, i.e., output coefficients of low pass filters. High pass coefficients are discarded since they provide detail information which serves no practical use for our application. Various levels of DWT are used to reduce the number of coefficients.

### III. IMPLEMENTATION STEPS

The image size used in the project work is 128 × 128 pixels. On applying the wavelet transform, the image is divided into approximate coefficients and detail coefficients. Level 1 yields the number of approximate coefficients as 64 × 64 = 4096. The approximate coefficients (low frequency coefficients) are stored and the detail coefficients (high frequency coefficients) are discarded. These approximate coefficients are used as inputs to the next level. Level 2 yields the number of approximate coefficients as 32 × 32 = 1024. These steps are repeated until an improvement in the recognition rate is observed. At each level the detail coefficients are neglected and the approximate coefficients are used as inputs to the next level. These approximate coefficients of input image and registered image are extracted. Each set coefficients belonging to the test image is compared with those of the registered image by taking the Euclidean distance and the recognition rate is calculated. Table 1 shows the comparison carried out at each level and its recognition rate.

TABLE I
COMPARISON OF VARIOUS LEVELS OF DWT

| Levels | Coefficients | Recognition Rate without normalized image | Recognition Rate with normalized image |
|---|---|---|---|
| Level 1 | 4096 | 85.1% | 91.4% |
| Level 2 | 1024 | 91.4% | 91.4% |
| Level 3 | 256 | 93.6% | 95.7% |
| Level 4 | 64 | 89% | 93.6% |
| Level 5 | 16 | 87.6% | 93.6% |

Upon inspecting the results obtained, we can infer that Level 3 offers better performance in comparison to other levels. Hence the images are subjected to decomposition only up to Level 3.

### IV. FUTURE WORK

The proposed face detection algorithm is time-efficient, i.e., having an execution speed of less than 1.75 seconds on an Intel Core 2 Duo 2.2 GHz processor. Due to its speed and robustness, it can further be extended for real time face detection and identification in video systems. Also, the proposed face recognition method can be coupled with recognition using local features, thus leading to an improvement in accuracy.

### V. CONCLUSION

The proposed face detection and extraction scheme was able to successfully extract the frontal face for poor resolutions as low as 320 × 200, even when the original image was poorly illuminated or extremely grainy. Thus, images obtained from low cost equipment like CCTV's and low resolution webcams could be processed by the algorithm.

For face recognition, the recognition rate for global features using various levels of DWT was calculated. Generally, the recognition rate was found to improve upon normalization. Level 3 DWT Decomposition gives a superior recognition rate as compared to other decomposition levels.

REFERENCES

[1] Z. Hafed, "Face Recognition Using DCT", *International Journal of Computer Vision*, 2001, pp. 167-188.
[2] W. Zhao, R. Chellappa, "Face Recognition: A Literature Survey", *ACM Computing Surveys*, Vol.35, No.4, December 2003, pp. 399-458, p. 9.

[3] S. Baker, T. Kanade, "Limits on super-resolution and how to break them," *IEEE Transactions on Pattern Analysis and Machine Intelligence*, Vol. 24, No. 9, pp. 1167-1183, September 2002.

[4] A. Braun, I. Jarudi and P. Sinha, "Face Recognition as a Function of Image Resolution and Viewing Distance," *Journal of Vision,* September 23, 2011, Vol. 11 No. 11, Article 666.

[5] L. S. Sayana, M. Tech Dissertation, *Face Detection*, Indian Institute of Technology (IIT) Bombay, p. 5, pp. 10-15.

[6] C. Schneider, N. Esau, L. Kleinjohann, B. Kleinjohann, "Feature based Face Localization and Recognition on Mobile Devices," *Intl. Conf. on Control, Automation, Robotics and Vision*, Dec. 2006, pp. 1-6.

[7] L. V. Praseeda, S. Kumar, D. S. Vidyadharan, "Face detection and localization of facial features in still and video images", *IEEE Intl. Conf. on Emerging Trends in Engineering and Technology*, 2008,pp.1-2.

[8] K. Chung, S. C. Kee, S. R. Kim, "Face Recognition Using Principal Component Analysis of Gabor Filter Responses," *International Workshop on Recognition, Analysis, and Tracking of Faces and Gestures in Real-Time Systems*, 1999, p. 53.

[9] T. Kawanishi, T. Kurozumi, K. Kashino, S. Takagi, "A Fast Template Matching Algorithm with Adaptive Skipping Using Inner-Subtemplates' Distances," Vol. 3, *17th International Conference on Pattern Recognition*, 2004, pp. 654-65.

[10] D. Margulis, *Photoshop Lab Colour: The Canyon Conundrum and Other Adventures in the Most Powerful Colourspace*, Pearson Education. ISBN 0321356780, 2006.

[11] J. Cai, A. Goshtasby, and C. Yu, "Detecting human faces in colour images," *Image and Vision Computing*, Vol. 18, No. 1, 1999, pp. 63-75.

[12] Singh, D. Garg, *Soft computing*, Allied Publishers, 2005, p. 222.

[13] S. Jayaraman, S. Esakkirajan, T. Veerakumar, *Digital Image Processing*, Mc Graw Hill, 2008.

[14] S. Assegie, M.S. thesis, Department of Electrical and Computer Engineering, Purdue University, *Efficient and Secure Image and Video Processing and Transmission in Wireless Sensor Networks*, pp. 5-7.

[15] P. Goyal, "NUC algorithm by calculating the corresponding statistics of the decomposed signal" , *International Journal on Computer Science and Technology (IJCST)*, Vol. 1, Issue 2, pp. 1-2, December 2010 .

[16] Y. Ma, C. Liu, H. Sun, "A Simple Transform Method in the Field of Image Processing", *Proceedings of the Sixth International Conference on Intelligent Systems Design and Applications*, 2006, pp. 1-2.

[17] H. A. Rowley, S. Baluja and T. Kanade, "Neural network-based face detection," *IEEE Transactions on Pattern Analysis and Machine Intelligence*, Vol. 20, No. 1, pp. 23-38, January 1998.

[18] P. Latha, L. Ganesan and S. Annadurai, "Face Recognition Using Neural Networks," *Signal Processing: An International Journal (SPIJ)*, Volume: 3 Issue: 5, pp. 153-160.

[19] S. Park, M. Park and M. Kang, "Super-resolution image reconstruction: a technical overview," *Signal Processing Magazine*, pp. 21-36, May 2003.

[20] P. Hennings-Yeomans, S. Baker, and B.V.K. Vijaya Kumar, "Recognition of Low-Resolution Faces Using Multiple Still Images and Multiple Cameras," *Proceedings of the IEEE International Conference on Biometrics: Theory, Systems, and Applications*, pp. 1-6, September 2008.

[21] MIT-CBCL Face Recognition Database, Center for Biological & Computational Learning (CBCL), Massachusetts Institute of Technology, Available: http://cbcl.mit.edu/software-datasets/heisele/facerecognition-database.html, July 2011.

[22] Multi-PIE Database, Carnegie Mellon University, Available: http://www.multipie.org, July 2011.

[23] The Yale Face Database, Department of Computer Science, Yale University, Available: http://cvc.yale.edu/projects/yalefacesB/yalefacesB.html, June 2011.

[24] T. Frajka, K. Zeger, "Downsampling dependent upsampling of images," *Signal Processing: Image Communication*, Vol. 19, No. 3, pp. 257-265, March 2004.

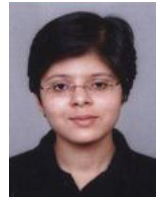
**Divya P. Jyoti** (M'2008) was born in Bhopal (M.P.) in India on April 24, 1990. She is currently pursuing her undergraduate studies in the Electronics and Telecommunication Engineering discipline at Thadomal Shahani Engineering College, Mumbai. Her fields of interest include Image Processing, and Human-Computer Interaction. She has 4 papers in International Conferences and Journals to her credit.

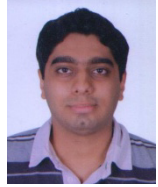
**Aman R. Chadha** (M'2008) was born in Mumbai (M.H.) in India on November 22, 1990. He is currently pursuing his undergraduate studies in the Electronics and Telecommunication Engineering discipline at Thadomal Shahani Engineering College, Mumbai. His special fields of interest include Image Processing, Computer Vision (particularly, Pattern Recognition) and Embedded Systems. He has 4 papers in International Conferences and Journals to his credit.

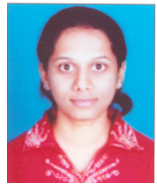
**Pallavi P. Vaidya** (M'2006) was born in Mumbai (M.H.) in India on March 18, 1985. She graduated with a B.E. in Electronics & Telecommunication Engineering from Maharashtra Institute of Technology (M.I.T.), Pune in 2006, and completed her post-graduation (M.E.) in Electronics & Telecommunication Engineering from Thadomal Shahani Engineering College (TSEC), Mumbai University in 2008. She is currently working as a Senior Engineer at a premier shipyard construction firm. Her special fields of interest include Image Processing and Biometrics.

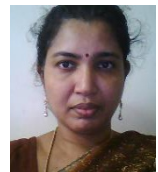
**M. Mani Roja** (M'1990) was born in Tirunelveli (T.N.) in India on June 19, 1969. She has received B.E. in Electronics & Communication Engineering from GCE Tirunelveli, Madurai Kamraj University in 1990, and M.E. in Electronics from Mumbai University in 2002. Her employment experience includes 21 years as an educationist at Thadomal Shahani Engineering College (TSEC), Mumbai University. She holds the post of an Associate Professor in TSEC. Her special fields of interest include Image Processing and Data Encryption. She has over 20 papers in National / International Conferences and Journals to her credit. She is a member of IETE, ISTE, IACSIT and ACM.